\documentclass[acmtog]{acmart}

\usepackage{multirow}
\usepackage{array}

\copyrightyear{2026}
\acmYear{2026}
\setcopyright{cc}
\setcctype{by}
\acmConference[SA Conference Papers '26]{SIGGRAPH Asia 2026 Conference Papers}{December 01--04, 2026}{Kuala Lumpur, Malaysia}
\acmBooktitle{SIGGRAPH Asia 2026 Conference Papers (SA Conference Papers '26), December 01--04, 2026, Kuala Lumpur, Malaysia}
\acmDOI{10.1145/3829340.3842375}
\acmISBN{979-8-4007-2842-6/2026/12}

\begin{document}

\title[HarmonyText: Harmonious Scene Text Editing via Reverse-to-Refined Pair Synthesis and Region-Decoupled Frequency Learning]{HarmonyText: Harmonious Scene Text Editing via Reverse-to-Refined Pair Synthesis and Region-Decoupled Frequency Learning}

\author{Ziqi Yi}
\authornote{Work done during an internship at Kolors Team, KlingAI Research.}
\orcid{0009-0002-2928-6746}
\affiliation{%
  \institution{South China University of Technology}
  \city{Guangzhou}
  \country{China}}
\affiliation{%
  \institution{Kolors Team, KlingAI Research}
  \city{Beijing}
  \country{China}}
\email{202421012296@mail.scut.edu.cn}

\author{Huiguo He}
\orcid{0000-0003-1419-059X}
\affiliation{%
  \institution{South China University of Technology}
  \city{Guangzhou}
  \country{China}}
\email{hehuiguo@scut.edu.cn}

\author{Guitao Xu}
\authornotemark[1]
\orcid{0009-0009-3023-4186}
\affiliation{%
  \institution{South China University of Technology}
  \city{Guangzhou}
  \country{China}}
\affiliation{%
  \institution{Kolors Team, KlingAI Research}
  \city{Beijing}
  \country{China}}
\email{eegtxu@mail.scut.edu.cn}

\author{Peirong Zhang}
\authornotemark[1]
\orcid{0000-0002-1857-5473}
\affiliation{%
  \institution{South China University of Technology}
  \city{Guangzhou}
  \country{China}}
\email{eeprzhang@mail.scut.edu.cn}

\author{Cheng Da}
\orcid{0000-0003-3037-7135}
\affiliation{%
  \institution{Kolors Team, KlingAI Research}
  \city{Beijing}
  \country{China}}
\email{dcfucheng@hotmail.com}

\author{Xinyao Yu}
\orcid{0009-0004-2499-8992}
\affiliation{%
  \institution{Kolors Team, KlingAI Research}
  \city{Beijing}
  \country{China}}
\email{yuxinyao22@mails.ucas.ac.cn}

\author{Zhanpeng Wang}
\orcid{0009-0004-5764-8577}
\affiliation{%
  \institution{Kolors Team, KlingAI Research}
  \city{Beijing}
  \country{China}}
\email{taotaow2023@gmail.com}

\author{Changqian Yu}
\orcid{0000-0002-4488-4157}
\affiliation{%
  \institution{Kolors Team, KlingAI Research}
  \city{Beijing}
  \country{China}}
\email{yuchangqian@kuaishou.com}

\author{Huan Yang}
\orcid{0009-0007-3756-1214}
\affiliation{%
  \institution{Kolors Team, KlingAI Research}
  \city{Beijing}
  \country{China}}
\email{hyang@fastmail.com}

\author{Kun Gai}
\orcid{0000-0002-3636-3618}
\affiliation{%
  \institution{Kolors Team, KlingAI Research}
  \city{Beijing}
  \country{China}}
\email{yuyue06@kuaishou.com}

\author{Lianwen Jin}
\orcid{0000-0002-5456-0957}
\authornote{Corresponding author.}
\affiliation{%
  \institution{South China University of Technology}
  \city{Guangzhou}
  \country{China}}
\email{eelwjin@scut.edu.cn}

\renewcommand{\shortauthors}{Yi et al.}

\begin{abstract}
Natural language-guided, mask-free scene text editing aims to modify text in real-world images using only textual instructions, yet remains challenged by unreliable pseudo-GT supervision, over-smoothed text generation, and the absence of GT-free evaluation. Existing methods often rely on imperfect pseudo-targets with background discrepancies, while standard MSE-based velocity regression tends to suppress high-frequency details, leading to blurred text boundaries and weakened local textures. To address these issues, we propose \textbf{HarmonyText}, a mask-free scene text editing framework with region- and frequency-aware supervision. HarmonyText introduces \textbf{Region-Decoupling Optimization (RDO)}, which aligns edited text regions with pseudo-targets while anchoring preserved regions to the source image, and \textbf{Cascaded High-Frequency Texture Guidance (CHFTG)}, which applies cascaded wavelet-domain constraints to preserve sharp boundaries and fine-grained textures of textual content. We further propose Reverse-to-Refined Pair Synthesis (RePair) to construct STE25K, a high-quality 25K paired dataset, and introduce a GT-free evaluation protocol with Color Distribution Divergence and a Harmony Assessment Model. Extensive experiments show that HarmonyText outperforms existing methods in text editing accuracy, background fidelity, color consistency, and perceptual harmony.
\end{abstract}



\maketitle


\begin{figure}[t]
    \centering
    \includegraphics[width=1.0\columnwidth]{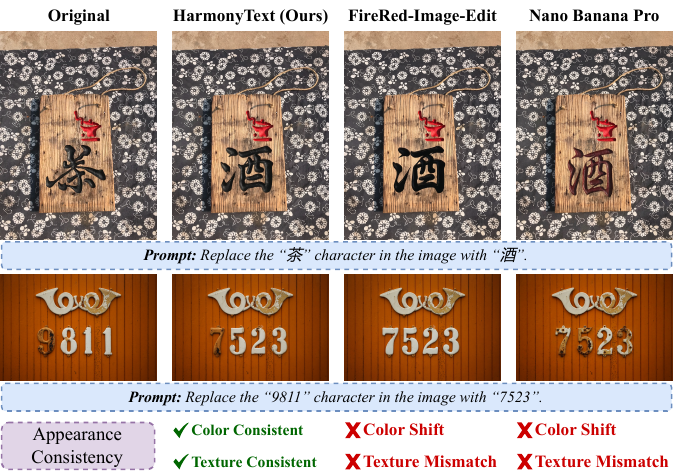}
    \caption{\textbf{Harmonious scene text editing.} Given only a natural-language instruction, HarmonyText edits the target text while preserving its natural integration with the surrounding scene, including local color and texture. In contrast, strong instruction-following editors may produce the correct text but introduce color shifts or texture mismatches. High-quality STE requires text accuracy together with visual harmony between the edited text and the scene.}
    \Description{Side-by-side scene text editing examples comparing HarmonyText with strong instruction-following editors. HarmonyText preserves the original scene appearance while changing the requested text, with more consistent local color, texture, and lighting.}
    \label{fig:teaser}
\end{figure}

\section{Introduction}
Scene Text Editing (STE) \cite{qu2023exploring, roy2020stefann, wu2019editing} aims to\let\thefootnote\relax\footnotetext{Project Lead: CHANGQIAN YU} modify textual content in natural images. This task has substantial practical value in applications such as poster design~\cite{zhao2025utdesign}, data generation for text recognition~\cite{zdenek2025otr}, and content-aware image processing~\cite{yu2025skyreels}.
It remains challenging because it requires accurate text generation while preserving the original background and ensuring that the edited text integrates naturally with its surroundings in color, texture, and lighting~\cite{zhang2025ocrgenbench}.

Early STE methods predominantly relied on explicit spatial masks or pre-rendered glyphs as auxiliary guidance~\cite{chen2023diffute, tuo2023anytext, xie2026textflux}. These priors provide strong control over text placement, but they require additional annotation effort and constrain editing flexibility. Recently, instruction-following generative models have enabled mask-free STE that operates with pure natural-language instructions~\cite{batifol2025flux, wu2025qwen, cui2025emu3, team2026firered}. However, as shown in Fig.~\ref{fig:teaser}, existing mask-free methods still suffer from noticeable color shifts, poor text-scene harmonization, and unintended background degradation, which compromise the realism and visual fidelity of the edited images.

A major obstacle in real-world STE is the lack of ground-truth (GT) editing pairs, i.e., paired images of the same scene that differ only in text content. This absence undermines both training supervision and real-world evaluation. (1) In model training, since real paired data cannot be collected at scale, existing methods typically rely on synthetic pseudo-GTs. However, such GTs often contain color shifts, texture inconsistencies, and background mismatches, making uniform pixel-wise supervision unreliable. Losses applied over the entire image may force the model to fit irrelevant discrepancies in unedited regions, degrading background fidelity. Moreover, spatial-domain objectives such as MSE tend to favor low-frequency reconstruction and suppress high-frequency details essential for realistic text rendering~\cite{wang2025analytical}. (2) For evaluation, reference-based metrics such as PSNR and SSIM are inapplicable without reliable GTs, while existing GT-free metrics are largely OCR-based and mainly measure text correctness, overlooking perceptual factors such as color consistency and text-background coherence.

To address these challenges, we propose \textbf{HarmonyText}, a novel framework for mask-free STE. First, to handle unreliable supervision from imperfect pseudo-GTs, HarmonyText introduces \textbf{Region-Decoupling Optimization (RDO)}. Specifically, RDO anchors preserved regions to the input image while aligning edited text regions with the pseudo-GT, thereby preventing background discrepancies in pseudo-GTs from being learned as supervision noise. Building on this region-aware formulation, we further introduce \textbf{Cascaded High-Frequency Texture Guidance (CHFTG)} to compensate for the low-frequency bias of MSE-based velocity regression. CHFTG applies a cascaded wavelet decomposition to high-frequency subbands and imposes region-aware frequency-domain supervision, encouraging sharper text boundaries and more faithful local textures in edited regions. Second, to enable GT-free evaluation, we introduce two complementary metrics: \textbf{Color Distribution Divergence (CDD)}, which measures color consistency between source and edited text, and a \textbf{Harmony Assessment Model}, which assesses text-background coherence. Third, to alleviate the scarcity of large-scale paired STE data, we propose a Reverse-to-Refined Pair Synthesis (RePair) pipeline. Instead of directly treating imperfect edits from existing models as pseudo-GTs, RePair first uses these flawed edited images as inputs and the corresponding real images as supervision targets, enabling the model to learn how realistic and well-integrated scene text should appear. The refined model is then used to generate higher-quality forward editing pairs, which are further filtered by CDD to construct \textbf{STE25K}, a high-quality STE dataset with 25K paired samples.

In summary, our contributions are as follows: 
\begin{itemize}
    \item We propose \textbf{HarmonyText}, a novel mask-free STE framework with two novel designs: \textbf{Region-Decoupling Optimization (RDO)} and \textbf{Cascaded High-Frequency Texture Guidance (CHFTG)}. RDO decouples edited and preserved regions with distinct supervision targets, while CHFTG imposes cascaded frequency-domain constraints to preserve sharp text boundaries and fine-grained textures.

    \item We propose two GT-free STE evaluation metrics, \textbf{Color Distribution Divergence (CDD)} and the \textbf{Harmony Score} produced by a \textbf{Harmony Assessment Model (HAM)}, which assess text color consistency and text-background coherence beyond OCR-based correctness.

    \item We introduce Reverse-to-Refined Pair Synthesis (RePair) to acquire large-scale STE data. By reversing imperfect edits toward real images and then using the refined model for forward synthesis, RePair produces higher-quality editing pairs. Based on RePair, we construct \textbf{STE25K}, a high-quality STE dataset containing 25K paired samples.

    \item Extensive experiments demonstrate that HarmonyText consistently outperforms existing methods across all evaluated dimensions, achieving superior background fidelity, color consistency, and text-background harmony simultaneously. We further assess generalization beyond HarmonyText-Bench on 345 images from six public scene text recognition datasets.
 
\end{itemize}

\section{Related Work}
\subsection{Scene Text Editing}

Early explorations in scene text editing (STE) predominantly relied on explicit spatial priors, such as editing region masks or pre-rendered text glyphs \cite{chen2023diffute, qu2023exploring, tuo2023anytext, fang2025cvpr, xie2026textflux, zhang2026weedit, zhao2024udifftext}. To enhance structural fidelity within these regions, mask-based methods such as UM-Text \cite{ma2026text} further introduce localized constraints like mask-guided Regional Consistency Loss. While effective for text placement and shape control \cite{lggpt2025zhang}, these hard priors require extra annotations or auxiliary inputs, limiting user flexibility and often leading to rigid layouts or unnatural blending.

Recently, the field has shifted toward instruction-guided, mask-free editing \cite{batifol2025flux, cui2025emu3, team2025longcat, wu2025qwen, omnigen22025wu, santoso2024on}. To compensate for the lack of explicit spatial priors, methods such as FireRed-Image-Edit \cite{team2026firered} introduce auxiliary consistency constraints beyond global spatial-domain losses. However, they struggle to selectively enhance text details while preserving untouched background regions, which often results in over-smoothed strokes, sticker-like artifacts, and degraded background textures. Frequency-domain control provides a promising way to address this limitation by modeling fine-grained visual details \cite{liao2025freqedit, jia2026icassp, yan2026glyphbanana}. For example, FreqEdit \cite{liao2025freqedit} leverages wavelet-based high-frequency feature injection to prevent texture collapse in multi-turn image editing, while GlyphBanana \cite{yan2026glyphbanana} employs latent-space frequency decomposition to inject high-frequency glyph details for text rendering. Nevertheless, existing frequency-domain methods rely on explicit masks and pre-rendered glyph priors, and are not specifically designed for mask-free STE. This motivates us to develop a frequency-aware editing mechanism that enhances texture preservation in text editing regions while also preserving the background details for the mask-free scenarios.

\subsection{Data Synthesis and Evaluation for STE}
High-quality training for STE requires paired before-and-after data from the same scene. Since manually collecting well-aligned real-world pairs is difficult, existing methods often rely on synthetic data construction or low-resource adaptation strategies \cite{seo2025stellar, zeng2024textctrl}. However, overlay-based synthesis can introduce artifacts and distribution biases, limiting realism and generalization. Recent studies have explored inverse data construction paradigms, where real images are treated as target outputs and corresponding inputs are derived through instruction inversion \cite{alekseenko2026vibe, xu2026towards}. This offers a more realistic alternative, but their potential for STE remains insufficiently explored.

Furthermore, evaluation of STE remains challenging. Common metrics such as PSNR, SSIM, and FID mainly measure global similarity or distribution-level quality, and are poorly suited to localized text edits. They often fail to capture local physical consistency, including color, texture, and text-background coherence. Recent benchmarks such as TextEditBench \cite{gui2025editbench} and REDEdit-Bench \cite{team2026firered} incorporate Large Language Model-based assessment, but such scoring can be unstable and still lacks dedicated objective measures of local physical realism.

\begin{figure*}[t]
    \centering
    \includegraphics[width=1.0\textwidth]{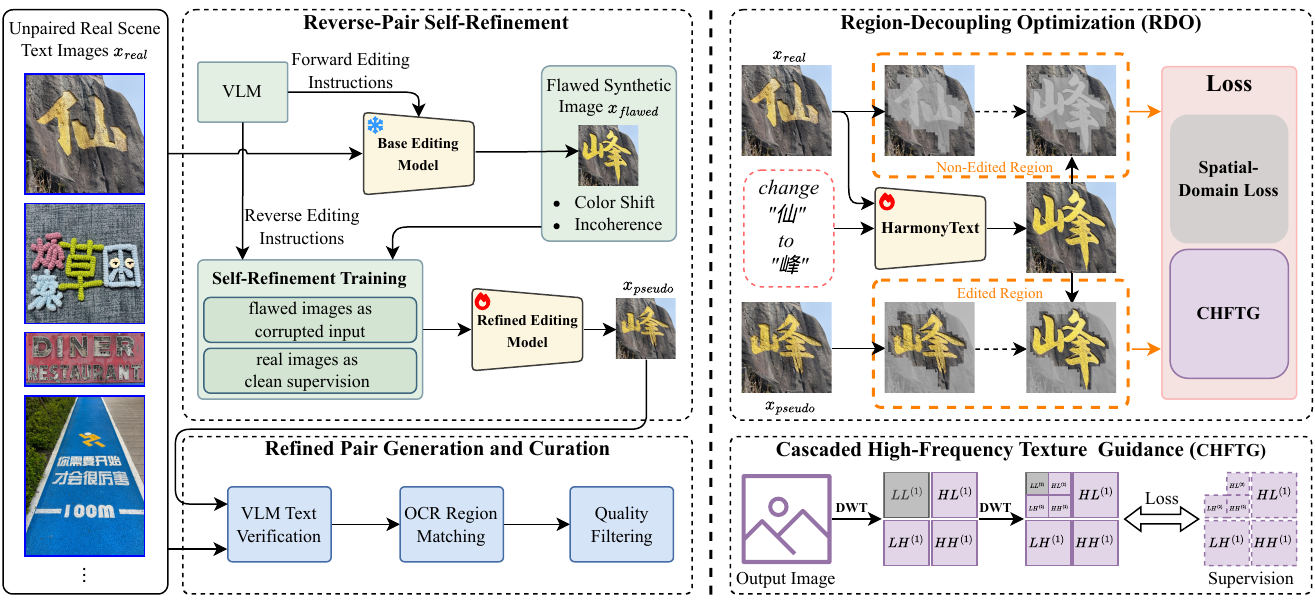}
    \caption{\textbf{Overview of the proposed HarmonyText framework.} (Left) RePair constructs high-quality paired data from unpaired real scene text images. A base editing model first produces flawed edits with color shifts and incoherence. The editing direction is then reversed, treating flawed outputs as corrupted inputs and original real images as clean supervision targets, training the model to reconstruct artifact-free scene text. The resulting pairs are further curated via text verification, region matching, and quality filtering to form STE25K. (Right) Region-Decoupled Optimization (RDO) assigns distinct objectives to edited and non-edited regions: non-edited regions are anchored to the source image for background preservation, while edited regions are supervised against the pseudo-target for accurate text generation. Cascaded High-Frequency Texture Guidance (CHFTG) further applies region-decoupled wavelet constraints across two frequency levels, preserving fine-grained textures and enforcing crisp text strokes.}
    \Description{Overview of HarmonyText. The left side shows the RePair data synthesis pipeline from real scene-text images through flawed edits, reverse self-refinement, forward generation, verification, region matching, and filtering. The right side shows Region-Decoupling Optimization and Cascaded High-Frequency Texture Guidance for edited and preserved regions.}
    \label{fig:framework}
\end{figure*}

\section{Method}

Our framework addresses the three core challenges of mask-free scene text editing: optimization misalignment, evaluation gaps, and data scarcity. Section~\ref{sec:harmonytext} details the HarmonyText model. Section~\ref{sec:eval} defines two fine-grained visual metrics for GT-free evaluation. Finally, Section~\ref{sec:ste25k} presents the RePair data synthesis pipeline and the resulting STE25K dataset.

\subsection{HarmonyText}
\label{sec:harmonytext}
To address the unreliable supervision caused by pseudo-target discrepancies and the over-smoothing tendency of standard MSE-based velocity regression, we propose \textbf{HarmonyText}, a mask-free scene text editing framework built upon region-aware flow matching, as illustrated in Fig.~\ref{fig:framework} (right). HarmonyText introduces two complementary components: \textbf{Region-Decoupling Optimization (RDO)}, which decouples edited and preserved regions with distinct supervision targets, and \textbf{Cascaded High-Frequency Texture Guidance (CHFTG)}, which imposes region-aware frequency-domain constraints to preserve sharp text boundaries and fine-grained textures.

\subsubsection{Region-Decoupling Optimization (RDO)}
Since the pseudo-target \(x_{pseudo}\) inevitably exhibits subtle background discrepancies from the source \(x_{real}\), we decouple edited and non-edited regions in latent space. This allows the model to align text regions with \(x_{pseudo}\) while anchoring background regions to \(x_{real}\) for faithful texture preservation.

Specifically, we employ the VAE encoder to extract the source latent \(z_s\) from \(x_{real}\) and the target latent \(z_t\) from \(x_{pseudo}\), respectively. We then compute the pixel-wise $L_{2}$ distance between the two latent representations along the channel dimension to obtain an edit-aware difference map $M_{diff}=\|z_t-z_s\|_2$. Since the VAE latent space preserves spatial and semantic information of the input image, non-edited regions are expected to have similar latent representations and thus small differences, while edited regions tend to produce larger deviations. Therefore, this difference map can serve as a soft cue to distinguish edited regions from non-edited regions.

Subsequently, we apply Min-Max normalization to \(M_{diff}\) to rescale its values into \([0,1]\), obtaining a soft edit-region mask \(M_{edit}\). To further enhance the contrast between edited and non-edited regions, we transform the normalized soft mask to the region-aware weighting maps $w_{edit}$ and $w_{pres}$ with an exponential weighting function:
\begin{equation}
w_{edit}=\alpha\cdot(\exp(k\cdot M_{edit})-1),
\end{equation}
\begin{equation}
w_{pres}=\alpha\cdot(\exp(k\cdot(1-M_{edit}))-1),
\end{equation}
where \(\alpha\) and \(k\) are scaling factors that control the overall weight magnitude and the contrast between regions, respectively. In our experiments, we empirically set \(\alpha=1.6\) and \(k=2.0\). This amplifies high-confidence edited regions in \(w_{edit}\) and high-confidence preserved regions in \(w_{pres}\), thereby strengthening the separation between generation and preservation objectives.

With \(w_{edit}\) and \(w_{pres}\), we adaptively decouple the optimization objectives for edited and preserved regions during Flow Matching training. Specifically, the edited regions are encouraged to follow the target velocity field \(v_t\) associated with the pseudo-target image \(x_{pseudo}\), enabling accurate text generation. In contrast, the preserved regions are constrained to follow the source-anchored velocity field \(v_s\) associated with the source image \(x_{real}\), thereby maintaining faithful background textures. The objectives are defined as:
\begin{align}
\mathcal{L}_{edit}
= \mathbb{E}_{(w_{edit}, \hat{v}, v_t)}\left[
\left\| w_{edit}\odot(\hat{v}-v_t) \right\|_2^2
\right],
\end{align}
\begin{align}
\mathcal{L}_{pres}
= \mathbb{E}_{(w_{pres}, \hat{v}, v_s)}\left[
\left\| w_{pres}\odot(\hat{v}-v_s) \right\|_2^2
\right],
\end{align}
where \(\odot\) denotes element-wise multiplication.

\subsubsection{Cascaded High-Frequency Texture Guidance (CHFTG)}
Although Region-Decoupling Optimization alleviates the issue of imperfect supervision caused by pseudo-target discrepancies, the standard MSE-based velocity regression in Flow Matching remains biased toward low-frequency structures, leading to blurred text edges and the loss of fine-grained textual content textures. To better preserve high-frequency textures, we introduce Cascaded High-Frequency Texture Guidance (CHFTG), which applies a cascaded wavelet decomposition to high-frequency subbands and supervises their detailed components.

Specifically, inspired by frequency feature extraction in other text-related fields \cite{frequency2026yi,dolphin2024zhang}, we apply a cascaded Daubechies-4 (DB4) Discrete Wavelet Transform (DWT)~\cite{daubechies1988orthonormal} to \(\hat{v}\), \(v_t\), and \(v_s\). At each decomposition level \(l \in \{1, 2\}\), the transform produces three high-frequency subbands indexed by \(d \in \{\mathrm{LH}, \mathrm{HL}, \mathrm{HH}\}\), corresponding to horizontal, vertical, and diagonal details at different spatial resolutions. We denote the resulting high-frequency subbands as \(\hat{v}^{(l,d)}\), \(v_t^{(l,d)}\), and \(v_s^{(l,d)}\). We further impose region-decoupled frequency-domain constraints on these multi-scale subbands to formulate the high-frequency optimization objectives:

\begin{align}
\mathcal{L}_{f\_edit}=\sum_{l,d}\mathbb{E}\left[\|\hat{w}_{edit}^{(l)}\odot(\hat{v}^{(l,d)}-v_t^{(l,d)})\|_2^2\right],
\end{align}
\begin{align}
\mathcal{L}_{f\_pres}=\sum_{l,d}\mathbb{E}\left[\|\hat{w}_{pres}^{(l)}\odot(\hat{v}^{(l,d)}-v_s^{(l,d)})\|_2^2\right],
\end{align}
where \(\hat{w}_{edit}^{(l)}\) and \(\hat{w}_{pres}^{(l)}\) denote the region-aware weighting maps resized to the spatial resolution of decomposition level \(l\).

Finally, the overall training objective is formulated as a weighted combination of the spatial region-decoupled losses and the high-frequency losses:
\begin{align}
\mathcal{L}_{total}
=
\lambda_{1}\mathcal{L}_{edit}
+
\lambda_{2}\mathcal{L}_{pres}
+
\lambda_3\mathcal{L}_{f\_edit}
+
\lambda_4\mathcal{L}_{f\_pres},
\end{align}
where \(\lambda_1\), \(\lambda_2\), \(\lambda_3\), and \(\lambda_4\) are the balancing coefficients. In our experiments, we set \(\lambda_1=1.0\), \(\lambda_2=1.5\), \(\lambda_3=0.2\), and \(\lambda_4=0.5\).

\subsection{Fine-Grained Visual Evaluation for STE}
\label{sec:eval}

Existing metrics either evaluate global fidelity (PSNR, SSIM, LPIPS) or text correctness (OCR), while recent MLLM-based holistic scoring~\cite{team2026firered} tends to be dominated by the unchanged background and overlooks subtle local artifacts. To fill this gap, we introduce two complementary metrics that jointly quantify visual harmony: the \textbf{Color Distribution Divergence (CDD)} score, which isolates color-level consistency between the source and edited text, and the \textbf{Harmony} score via the proposed Harmony Assessment Model, which captures the broader perceptual integration including surface texture adherence and lighting coherence.

\subsubsection{Color Distribution Divergence}
\label{color}

Inspired by CVD Loss~\cite{zhuang2026prompt}, which accounts for structural degradation caused by color shifts, we propose the CDD score, a CIELAB-based statistical distribution divergence metric~\cite{robertson1978colorimetry} to quantify the color consistency between original and edited text.

Specifically, given the source image \(x_s\) and the edited image \(x_{pred}\), we first crop the corresponding local regions from both images using the bounding box of the edited region. A SOTA scene text segmentation method, HiSAM~\cite{HiSAM}, is then applied to the cropped regions to obtain localized binary text masks. Next, we convert the cropped source and edited regions from RGB to the CIELAB color space. Based on the text masks, we extract the corresponding CIELAB color values and compute normalized 1D color histograms independently for the \(L\), \(a\), and \(b\) channels. The resulting distributions are denoted as \(P_c\) and \(Q_c\), where \(c \in \{L,a,b\}\), representing the source and edited text color distributions, respectively. To quantify the color distributions discrepancy between the source and edited text, we adopt the Jensen-Shannon (JS) divergence, which is symmetric and bounded. For each channel, we define the mean distribution as \(M_c=\frac{1}{2}(P_c+Q_c)\). The CDD score is then computed as the average JS divergence over the three CIELAB channels:
\begin{align}
\mathrm{CDD}
=
\frac{1}{3}
\sum_{c\in\{L,a,b\}}
\left(
\frac{1}{2}D_{\mathrm{KL}}(P_c\parallel M_c)
+
\frac{1}{2}D_{\mathrm{KL}}(Q_c\parallel M_c)
\right),
\end{align}
where \(D_{\mathrm{KL}}(P \parallel Q)=\sum_{i=1}^{N}P(i)\ln\frac{P(i)}{Q(i)}\), and \(N\) denotes the number of histogram bins. A lower CDD score indicates better color consistency between the source and edited text.

\subsubsection{Harmony Score}
Existing metrics are insufficient for assessing whether edited text is naturally integrated into the surrounding scene, particularly in terms of text-background coherence and local texture fidelity. To address this, we build a \textbf{Human Harmony Reference (H2R)} dataset and train a \textbf{Harmony Assessment Model (HAM)} to predict a harmony score from the source and edited regions, measuring how well the edited text preserves local texture and blends with its background.

\paragraph{H2R Dataset}
We collect 80K scene text images and edit each with three models: Qwen-Image-Edit, FireRed-Image-Edit-v1.1, and our refined model from Section~\ref{sec:ste25k}, producing 240K edited samples in total. Each sample is manually rated on a five-point scale with reference to the source image. Annotators focus exclusively on the edited region and evaluate surface texture adherence, lighting coherence, and overall naturalness (detailed guidelines in the supplementary material); samples with failed text generation are removed during annotation. To construct region-level training pairs, we use PaddleOCR to localize edited regions and crop the corresponding source and edited patches. Qwen3-VL is further employed to filter out inaccurate localizations. Finally, we pair edited regions from different models under the same source image, yielding 52K region-level comparison samples for pairwise preference learning.

\paragraph{HAM as an Evaluator}
Human ratings of visual harmony are inherently subjective and may contain annotator variance and noise. To learn a robust evaluation model from such annotations, we train the evaluator HAM with an uncertainty-aware preference learning objective. Following EditReward~\cite{wu2025editreward}, HAM takes the source region, text prompt, and edited region as inputs. Qwen3-VL is used as the multimodal backbone to extract rich visual-semantic features, and an MLP head predicts a Gaussian harmony score distribution \(s\sim\mathcal{N}(\mu,\sigma^2)\), where \(\mu\) represents the expected harmony score and \(\sigma^2\) captures prediction uncertainty.

During training, we construct pairwise samples \((I_h,I_l)\) from edited regions generated for the same source image and prompt, where \(I_h\) is preferred over \(I_l\) according to human annotations. Following probabilistic ranking frameworks~\cite{ma2025hpsv3}, we compute the preference probability \(P(I_h\succ I_l)\) from their predicted score distributions and optimize HAM with:
\begin{align}
\mathcal{L}_{\mathrm{harmony}}
=
-\log P(I_h\succ I_l).
\end{align}
Notably, all training pairs are constructed from outputs of the above three models, excluding the final HarmonyText model. This ensures that HAM captures general human visual preferences rather than overfitting the output distribution of any particular model. During inference, the predicted mean \(\mu\) is used as the final harmony score.

\subsection{STE25K Dataset Construction}
\label{sec:ste25k}

Since STE pairs do not naturally exist in real-world scenarios, constructing high-quality STE datasets remains challenging. Existing methods typically simulate paired data by inserting text into natural scene images, which often results in poor text-background coherence. To address this issue, we introduce \textbf{STE25K}, a 25K-scale high-quality STE dataset constructed via our \textbf{Reverse-to-Refined Pair Synthesis (RePair)} pipeline, as shown in Fig.~\ref{fig:framework} (left). RePair comprises two stages: reverse-pair self-refinement and refined pair generation with filtering.

\subsubsection{Reverse-Pair Self-Refinement}
We collect over 100K real-world scene text images with diverse styles from the Internet. Given a scene text image \(x_{real}\), we employ Qwen3-VL to generate a corresponding text editing instruction \(t_{edit}\). The editing model $\mathcal{G}$ (e.g., FireRed-Image-Edit) edits \(x_{real}\) under the guidance of the text editing instruction \(t_{edit}\), producing an edited image \(x_{flawed}\) that often contains visual artifacts. We then reverse the editing direction to construct a reverse pair:
\begin{align}
x_{flawed} \xrightarrow{\,\tilde{t}_{edit}\,} x_{real},
\end{align}
where \(\tilde{t}_{edit}\) is the reverse editing instruction obtained by swapping the source and target text in \(t_{edit}\). Finally, we obtain more than 100K reverse pairs for self-refinement training.

Rather than using \(x_{flawed}\) as the pseudo ground truth, we train the model on the reverse pairs to reconstruct the original real image:
\begin{align}
\hat{x}_{real}=\mathcal{G}_\theta(x_{flawed}, \tilde{t}_{edit}),
\end{align}
with the optimization objective:
\begin{align}
\min_\theta \mathbb{E}_{(x_{flawed}, \tilde{t}_{edit}, x_{real})}\left[
\mathcal{D}\left(\hat{x}_{real}, x_{real}\right)
\right],
\end{align}
where $\mathcal{D}$ denotes the loss function. This self-refinement objective guides the model to map artifact-contaminated samples back toward the real image distribution, enabling it to acquire an artifact-to-real refinement prior.

\subsubsection{Refined Pair Generation and Filtering}
The refined model is then applied to \(x_{real}\) with the forward instruction \(t_{edit}\), producing a higher-quality edited result \(x_{pseudo}\). Together with \(x_{real}\) and \(t_{edit}\), this forms a refined STE pair:
\begin{align}
x_{real} \xrightarrow{\,t_{edit}\,} x_{pseudo}.
\end{align}

Initially, we obtain over 100K refined STE pairs. We then verify whether the text in \(x_{pseudo}\) matches the target text specified by \(t_{edit}\), and discard samples with incorrect text, yielding 80K instruction-following STE pairs.

Next, we obtain OCR results from both \(x_{real}\) and \(x_{pseudo}\) using PaddleOCR. We match the corresponding text instances before and after editing based on the detected bounding boxes and recognized text. The editing region \(M_{e}\) is defined as the union of the matched source and target text regions, with the non-edited region \(M_{bg} = 1 - M_{e}\). We then apply region-specific quality filtering. For background preservation, we compute PSNR and SSIM on \(M_{bg}\) in the VAE latent space. Because pseudo-GTs undergo VAE encoding and decoding, pixel-level comparisons are dominated by reconstruction noise; operating in latent space filters out this noise and better reflects true quality. For color consistency, we compute CDD (Section~\ref{color}) on \(M_{e}\). CDD is parameter-free, using standard JS divergence in CIELAB space without learned components, and thus provides a filtering criterion independent of model parameters. Applying metric-specific thresholds yields 25K high-quality STE pairs, denoted as \textbf{STE25K}. 

\begin{table}[t]
\centering
\caption{Harmony preference alignment accuracy on HarmonyText-Bench. Accuracy measures the percentage of pairwise preference judgments that agree with human annotations.}
\renewcommand{\arraystretch}{0.95}
\footnotesize
\begin{tabular}{@{} p{2.5cm} p{2.0cm} >{\centering\arraybackslash}p{1.5cm} @{}}
\toprule
\textbf{Category} & \textbf{Method} & \textbf{Accuracy} \\
\midrule
Baseline & Random & 49.80 \\
\midrule
\multirow{2}{*}{\textit{Commercial Models}} & GPT-5.4 & 53.88 \\
& Gemini-3.1-Pro & 55.29 \\
\midrule
\multirow{2}{*}{\textit{Open-Source VLMs}} & Qwen3.6-27B & 48.06 \\
& Qwen3-VL-8B & 35.68 \\
\midrule
& \textbf{Ours} & \textbf{83.07} \\
\bottomrule
\end{tabular}
\label{tab:genai_bench_comparison}
\end{table}

\begin{table*}[t]
\centering
\caption{Quantitative comparison on HarmonyText-Bench (155 self-curated images) and Public-STR (345 public-source images) using the original short instructions. Non-edited region metrics (PSNR, SSIM, LPIPS) assess background preservation, while edited region metrics (ACC, NED, CDD, Harm.) evaluate text accuracy, color consistency, and text-background coherence. Harm. denotes Harmony score. The best and second-best results within each dataset are highlighted in \textbf{bold} and with \underline{underlines}, respectively.}
\renewcommand{\arraystretch}{0.95}
\setlength{\tabcolsep}{1.2pt}
\resizebox{\textwidth}{!}{
\begin{tabular}{lc ccc cccc @{\hspace{4pt}} ccc cccc}
\toprule
\multirow{3}{*}{\textbf{Method}} & \multirow{3}{*}{\textbf{Mask-free}}
& \multicolumn{7}{c}{\textbf{HarmonyText-Bench (155)}}
& \multicolumn{7}{c}{\textbf{Public-STR (345)}} \\
\cmidrule(lr){3-9}\cmidrule(lr){10-16}
& & \multicolumn{3}{c}{\textbf{Non-Edited}} & \multicolumn{4}{c}{\textbf{Edited}}
& \multicolumn{3}{c}{\textbf{Non-Edited}} & \multicolumn{4}{c}{\textbf{Edited}} \\
\cmidrule(lr){3-5}\cmidrule(lr){6-9}\cmidrule(lr){10-12}\cmidrule(lr){13-16}
& & \textbf{PSNR\(\uparrow\)} & \textbf{SSIM\(\uparrow\)} & \textbf{LPIPS\(\downarrow\)} & \textbf{ACC\(\uparrow\)} & \textbf{NED\(\uparrow\)} & \textbf{CDD\(\downarrow\)} & \textbf{Harm.\(\uparrow\)}
& \textbf{PSNR\(\uparrow\)} & \textbf{SSIM\(\uparrow\)} & \textbf{LPIPS\(\downarrow\)} & \textbf{ACC\(\uparrow\)} & \textbf{NED\(\uparrow\)} & \textbf{CDD\(\downarrow\)} & \textbf{Harm.\(\uparrow\)} \\
\midrule
AnyText2 \cite{tuo2024anytext2} & \(\times\) & \underline{27.226} & \underline{0.783} & 0.190 & 0.548 & 0.711 & 0.385 & -2.967 & \underline{31.853} & \underline{0.891} & 0.127 & 0.736 & 0.880 & 0.420 & -2.970 \\
DreamText \cite{wang2025dreamtext} & \(\times\) & 23.164 & 0.646 & 0.224 & 0.181 & 0.360 & 0.356 & -2.333 & 25.868 & 0.813 & 0.158 & 0.043 & 0.174 & 0.353 & -2.251 \\
BAGEL \cite{deng2025emerging} & \(\checkmark\) & 25.390 & 0.782 & 0.148 & 0.406 & 0.668 & 0.269 & -1.717 & 28.572 & 0.883 & 0.125 & 0.232 & 0.581 & 0.284 & -1.410 \\
Step1X-Edit-Think \cite{liu2025step1x} & \(\checkmark\) & 23.169 & 0.670 & 0.152 & 0.761 & 0.921 & 0.175 & -0.207 & 25.289 & 0.805 & 0.115 & 0.707 & 0.913 & 0.178 & 0.347 \\
Qwen-Image-Edit \cite{wu2025qwen} & \(\checkmark\) & 20.611 & 0.673 & 0.154 & 0.832 & 0.958 & 0.180 & -2.996 & 22.793 & 0.815 & 0.134 & 0.878 & 0.966 & 0.175 & -2.832 \\
LongCat-Image-Edit \cite{team2025longcat} & \(\checkmark\) & 22.359 & 0.672 & 0.159 & 0.910 & 0.975 & 0.279 & -1.967 & 24.192 & 0.798 & 0.140 & 0.890 & 0.968 & 0.296 & -1.920 \\
Flux.2-dev \cite{flux22025} & \(\checkmark\) & 23.773 & 0.609 & 0.131 & 0.690 & 0.864 & 0.096 & 1.915 & 25.328 & 0.766 & 0.125 & 0.626 & 0.854 & \underline{0.075} & 2.278 \\
Flux.2-Klein-9B \cite{flux2klein2026} & \(\checkmark\) & 23.837 & 0.591 & 0.143 & 0.477 & 0.634 & 0.118 & 2.530 & 25.868 & 0.747 & 0.128 & 0.214 & 0.472 & 0.114 & 2.140 \\
FireRed-Image-Edit-v1.1 \cite{team2026firered} & \(\checkmark\) & 26.796 & 0.774 & 0.125 & 0.865 & 0.968 & 0.152 & 0.560 & 30.020 & 0.870 & 0.098 & 0.907 & 0.984 & 0.136 & 1.026 \\
\midrule
Nano Banana Pro \cite{google_gemini_nanobanana_2025} & \(\checkmark\) & 23.853 & 0.675 & \underline{0.114} & \underline{0.923} & 0.968 & \underline{0.078} & \underline{2.606} & 26.621 & 0.807 & \textbf{0.085} & 0.939 & 0.977 & 0.083 & \underline{2.512} \\
Seedream 4.5 \cite{seedream2025seedream} & \(\checkmark\) & 21.939 & 0.703 & \textbf{0.103} & 0.916 & 0.964 & 0.160 & 0.722 & 24.104 & 0.819 & 0.098 & 0.899 & 0.965 & 0.189 & 0.241 \\
GPT Image 2.0 \cite{gptimage22026} & \(\checkmark\) & 18.947 & 0.560 & 0.223 & \textbf{0.935} & \textbf{0.984} & 0.121 & 1.902 & 17.465 & 0.639 & 0.237 & \textbf{0.971} & \underline{0.985} & 0.193 & 0.943 \\
\midrule
\textbf{HarmonyText (Ours)} & \(\checkmark\) & \textbf{28.443} & \textbf{0.803} & \underline{0.114} & 0.916 & \underline{0.981} & \textbf{0.063} & \textbf{2.761} & \textbf{33.040} & \textbf{0.903} & \underline{0.089} & \underline{0.948} & \textbf{0.988} & \textbf{0.050} & \textbf{2.753} \\
\bottomrule
\end{tabular}
}
\label{tab:method_comparison}
\end{table*}

\section{Experiment}
\subsection{Implementation Details}
The self-refinement model and HarmonyText are initialized from FireRed-Image-Edit-v1.1~\cite{team2026firered} and trained with separate LoRA adapters applied only to the Transformer attention projections. Both adapters use a rank of 16 and a scaling factor of 32. The AdamW optimizer is adopted with \((\beta_1,\beta_2)=(0.9,0.999)\). The learning rate and weight decay are set to \(2\times10^{-4}\) and 0.01, respectively. The self-refinement adapter is trained for 2 epochs solely to generate refined forward pairs and is then discarded; HarmonyText uses a newly initialized adapter trained on STE25K for 5 epochs. Training takes approximately 20 hours on 8 GPUs with a per-GPU batch size of 1. RDO and CHFTG are training-only objectives with no additional inference modules or sampling steps. HAM is trained separately for 4 epochs, with its complete configuration in the supplementary material.

\subsection{Evaluation}
\paragraph{Benchmark}
To facilitate the comprehensive evaluation of STE, we introduce \textbf{HarmonyText-Bench}, a benchmark containing 155 high-resolution scene text images with diverse text styles and challenging backgrounds. These backgrounds include complex patterns such as stone textures, illumination variations, and fabric wrinkles, which impose high requirements on generating edited text that is naturally integrated into the surrounding scene. Each image is equipped with a manually annotated text editing prompt and a binary edit-region mask specifying the editing region. HarmonyText-Bench is disjoint from both STE25K and H2R at the source-image level, ensuring that no benchmark image is used to train either HarmonyText or HAM.

\paragraph{Public-STR Generalization Set}
To evaluate generalization beyond HarmonyText-Bench, we sample 345 high-resolution images from six public scene text recognition datasets: Total-Text~\cite{ch2017total}, SCUT-CTW1500~\cite{liu2019curved}, ArT~\cite{chng2019icdar2019}, TextOCR~\cite{singh2021textocr}, RCTW-17~\cite{shi2017icdar2017}, and ReCTS~\cite{zhang2019icdar}. Following the same annotation protocol, each image has a manually annotated editing prompt and binary edit-region mask. Together with the 155-image HarmonyText-Bench, this independently sourced subset forms a 500-image evaluation suite covering diverse text styles, backgrounds, and real-world scenarios.

\paragraph{Metrics}
Using the edit-region mask, we separately evaluate the edited and non-edited regions. For the non-edited region, we adopt standard fidelity metrics, including PSNR, SSIM, and LPIPS, to assess background preservation and texture consistency. For the edited region, we evaluate text correctness using accuracy (ACC) and normalized edit distance (NED), while using CDD and Harmony Score to assess color consistency and visual harmony.

\paragraph{User Study}
In a blinded A/B user study on HarmonyText-Bench, 8 participants (4 female and 4 male, aged 24--50) provide 1,240 judgments for each HarmonyText--baseline comparison. Method identities are hidden, presentation order and left/right positions are randomized, and ties are allowed. Participants jointly assess instruction following (spelling, color, and natural blending) and background preservation (degradation, over-smoothing, and artifacts). We report win and loss percentages, with the remainder representing ties.

\subsection{Validation of Proposed Metrics }

\begin{figure}[!t]
    \centering
    \includegraphics[width=\linewidth]{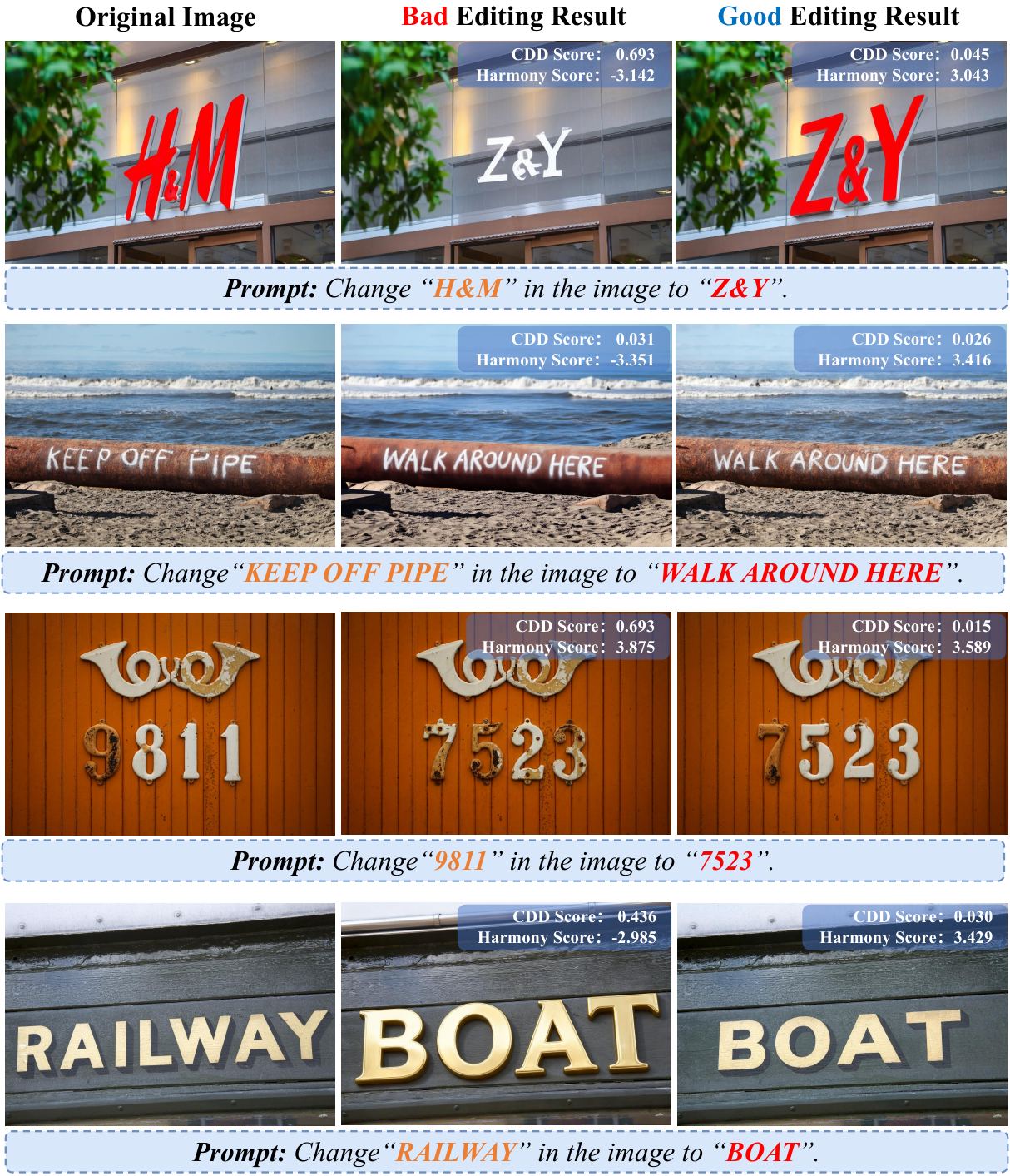}
    \caption{\textbf{Qualitative validation of the proposed CDD and Harmony scores.} For each example, we show the source image and edited results from different methods, along with the corresponding CDD and Harmony scores. A lower CDD indicates better color consistency: edited text with noticeable color drift or illumination mismatch yields higher CDD values, while naturally blended text yields lower ones. A higher Harmony score indicates better perceptual integration: it consistently assigns higher ratings to edits that exhibit better texture adherence and lighting coherence with the surrounding scene, aligning well with human perception.}
    \Description{Examples used to validate CDD and Harmony scores. For each source image, edited results from different methods are shown with their CDD and Harmony values, illustrating that lower CDD corresponds to better color consistency and higher Harmony scores correspond to more natural texture and lighting integration.}
    \label{fig:cdd-harm}
\end{figure}

As illustrated in Fig.~\ref{fig:cdd-harm}, CDD scores correlate well with visual perception: edited text suffering from color drift or illumination mismatch consistently yields higher JS divergence, while naturally blended text yields low CDD values. For HAM, we construct 1,494 human-annotated preference pairs and measure the agreement rate between HAM's predicted ranking and human judgment. As shown in Tab.~\ref{tab:genai_bench_comparison}, HAM achieves 83.07\% alignment accuracy, substantially outperforming commercial VLMs (GPT-5.4: 53.88\%, Gemini-3.1-Pro: 55.29\%) and open-source alternatives (Qwen3.6-27B: 48.06\%). This confirms that general-purpose VLMs lack the fine-grained sensitivity to local texture and color integration required for harmony evaluation. Notably, HAM generalizes well to models absent from its training data (e.g., Nano Banana Pro, LongCat-Image-Edit, Step1X-Edit-Think, and BAGEL in Tab.~\ref{tab:method_comparison}), indicating that it captures general harmony preferences rather than overfitting to specific model distributions. Together, CDD and Harmony Score provide complementary evaluation: CDD quantifies color-level consistency, while HAM captures texture and coherence at the perceptual level. The HAM training set excludes all final HarmonyText outputs, and HAM never serves as a HarmonyText training objective or data-selection signal. Our conclusions are further supported by the blinded user study on HarmonyText-Bench and the standard fidelity and recognition metrics on Public-STR, both of which are independent of CDD and HAM.

\subsection{Comparison with SOTA Methods}

\paragraph{Results on HarmonyText-Bench}
Tab.~\ref{tab:method_comparison} presents a comprehensive comparison of both mask-based and mask-free methods on HarmonyText-Bench. For background preservation, HarmonyText achieves the highest PSNR (28.443~dB) and SSIM (0.803) among all compared methods, surpassing the strongest open-source baseline FireRed-Image-Edit-v1.1 by 1.65~dB in PSNR. On the two proposed perceptual metrics, HarmonyText attains the lowest CDD (0.063), reducing it by 19.2\% relative to the next best method (Nano Banana Pro, 0.078), while achieving the highest Harmony Score (2.761). These gains indicate that the generated text closely preserves the color distribution of the original scene text and integrates naturally into the surrounding background. Meanwhile, HarmonyText maintains competitive text accuracy (ACC of 0.916, NED of 0.981). Although GPT Image 2.0 achieves marginally higher ACC (0.935), it suffers from severe background degradation (PSNR of only 18.947~dB) and substantially lower visual harmony. These results demonstrate that HarmonyText consistently outperforms existing methods across all evaluated dimensions, achieving superior background fidelity, color consistency, and text-background harmony simultaneously without sacrificing text generation accuracy.

\paragraph{Generalization on Public STR Datasets}
The Public-STR columns of Tab.~\ref{tab:method_comparison} report results on 345 images using the original short instructions for all methods. HarmonyText ranks first in PSNR (33.040), SSIM (0.903), NED (0.988), CDD (0.050), and Harmony Score (2.753), and second in LPIPS (0.089) and ACC (0.948). Its CDD is 33.3\% lower than the second-best result, showing that the color-consistency gain generalizes beyond HarmonyText-Bench. GPT Image 2.0 achieves the highest ACC but lower background fidelity and harmony, while Nano Banana Pro achieves the lowest LPIPS but trails HarmonyText on the other background- and harmony-oriented metrics. Overall, HarmonyText provides the best balance of background preservation, text correctness, and visual integration.

\begin{figure*}[!t]
    \centering
    \includegraphics[width=0.94\textwidth]{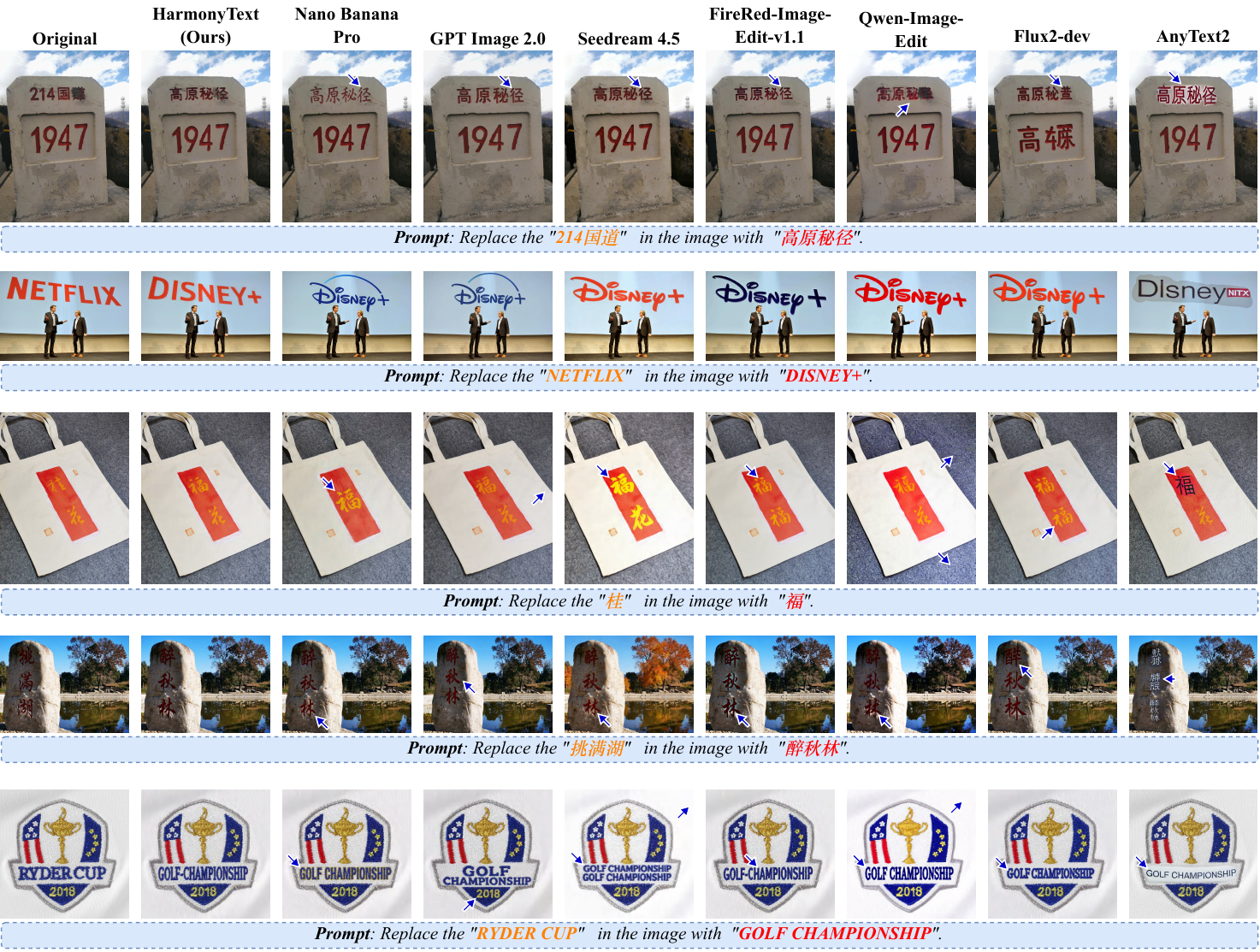}
    \caption{\textbf{Qualitative comparisons of scene text editing between our HarmonyText and other SOTA methods.} Arrows indicate regions where the edited result deviates from the original appearance in color, texture, or coherence. Our HarmonyText achieves the best visual harmony, with superior texture preservation, color consistency, and text-background coherence. \textbf{Best viewed with zoom-in.}}
    \Description{Qualitative comparison grid for several Chinese and English scene-text editing prompts. Each row shows the source image and outputs from multiple baselines and HarmonyText; arrows mark color, texture, or coherence artifacts. HarmonyText more closely preserves the source appearance while producing the requested text.}
    \label{fig:compare_sota}
\end{figure*}

\paragraph{Qualitative Comparison}
Fig.~\ref{fig:compare_sota} presents visual comparisons across diverse editing scenarios involving both Chinese and English text on various surfaces. HarmonyText consistently produces the most visually harmonious edits, preserving the original color, surface texture, and lighting while generating accurate text. In contrast, competing methods exhibit varying degrees of degradation. For instance, in the stone monument example (row 1), our method reproduces not only the carved text style but also subtle surface details such as weathering stains and natural discoloration, whereas other methods either alter the background scene or produce strokes that lack the authentic engraved appearance. Similar patterns hold across other examples. The user study in Tab.~\ref{tab:user_study} corroborates these observations: HarmonyText is preferred over all baselines, achieving win rates above 82\% against most methods and over 53\% even against the strongest commercial competitors (e.g., Nano Banana Pro). These results confirm the effectiveness of the proposed region-decoupled frequency optimization and RePair pipeline in jointly achieving faithful background preservation and photorealistic text integration.

\begin{table}[t]
\centering
\footnotesize
\renewcommand{\arraystretch}{1.0}
\caption{Blinded A/B user-study results on HarmonyText-Bench (\%). Each HarmonyText--baseline comparison contains 1,240 judgments; omitted percentages correspond to ties.}
\label{tab:user_study}
\begin{tabular*}{\linewidth}{@{\extracolsep{\fill}} l c c @{}}
\toprule
\multirow{2}{*}{Method} & \multicolumn{2}{c}{HarmonyText vs. Baseline} \\
\cmidrule(lr){2-3}
 & Win & Lose \\
\midrule
AnyText2~\cite{tuo2024anytext2}                 & \textbf{99.9} & 0.1 \\
DreamText~\cite{wang2025dreamtext}              & \textbf{100.0} & 0.0 \\
Qwen-Image-Edit~\cite{wu2025qwen}               & \textbf{96.5} & 2.7 \\
LongCat-Image-Edit~\cite{team2025longcat}       & \textbf{93.3} & 6.5 \\
Step1X-Edit-Think~\cite{liu2025step1x}          & \textbf{91.9} & 6.8 \\
Flux.2-dev~\cite{flux22025}                     & \textbf{82.6} & 15.1 \\
FireRed-Image-Edit-v1.1~\cite{team2026firered}  & \textbf{88.9} & 5.0 \\
Nano Banana Pro~\cite{google_gemini_nanobanana_2025} & \textbf{53.2} & 42.6 \\
Seedream 4.5~\cite{seedream2025seedream}        & \textbf{76.3} & 21.4 \\
GPT Image 2.0~\cite{gptimage22026}              & \textbf{61.7} & 37.5 \\
\bottomrule
\end{tabular*}
\end{table}

\subsection{Ablation Studies}
\paragraph{Ablation on RePair}
Tab.~\ref{tab:ablation_repair} evaluates the training strategies and stages of RePair, with all variants trained using the same HarmonyText model. To justify our final training direction, we introduce a Reverse Direction baseline that trains on the same STE25K pairs but in the reverse direction (edited $\to$ real). Although it improves upon the base model, it is significantly outperformed by forward training: its CDD nearly doubles (0.063$\to$0.121) and the Harmony score drops sharply. This performance gap arises because reverse-direction training cannot leverage our Region-Decoupled Frequency Learning. Using original real images as supervision targets enables the refinement model to learn an in-distribution prior for realistic text appearance before generating forward pairs, without explicit source-style optimization.

\begin{figure}[!t]
    \centering
    \includegraphics[width=\linewidth]{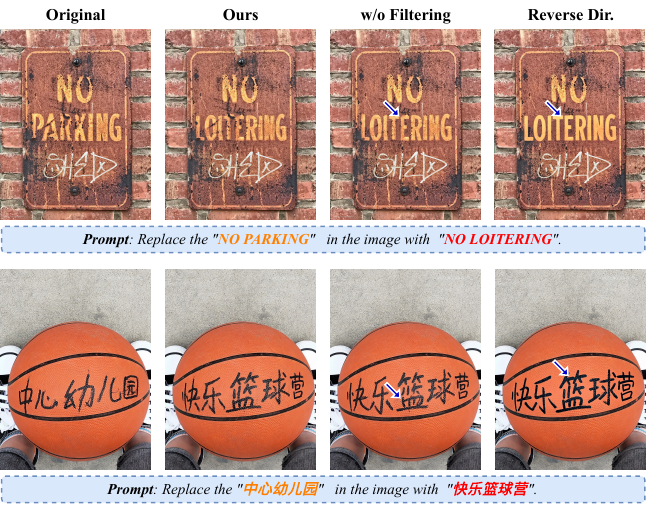}
    \caption{\textbf{Visual ablation on the RePair data synthesis pipeline.} \textit{Reverse Dir.} trains on STE25K pairs in the reverse direction (edited $\to$ real) instead of the forward direction. \textit{w/o Filtering} uses forward-direction pairs without quality filtering. Arrows indicate regions where color or texture deviates from the original image. The full RePair pipeline yields the best overall consistency. \textbf{Best viewed with zoom-in.}}
    \Description{Visual ablation of the RePair data synthesis pipeline comparing Reverse Direction, forward training without filtering, and the full RePair method. Arrows indicate regions with color or texture inconsistencies; the full pipeline shows the most consistent appearance.}
    \label{fig:ablation_data}
\end{figure}

Thus, forward training is adopted. Even without multi-metric filtering, the forward-trained model outperforms the Reverse Direction baseline by a large margin. The filtering provides further complementary gains (CDD from 0.067 to 0.063). As shown in Fig.~\ref{fig:ablation_data}, the Reverse Direction variant produces noticeable color and texture inconsistencies in the non-edited regions (indicated by blue arrows), while removing only data filtering leads to subtler but still visible artifacts. The full RePair pipeline achieves the most faithful background preservation and natural text integration.

\begin{table}[t]
\centering
\caption{Ablation study on the RePair. \textit{Reverse Direction} denotes a baseline that trains on the same STE25K pairs but in the reverse direction, taking the edited images as input to reconstruct the original real images. All variants are evaluated on HarmonyText-Bench.}
\label{tab:ablation_repair}
\renewcommand{\arraystretch}{0.90}
\resizebox{\linewidth}{!}{
\begin{tabular}{l|ccc|cccc}
\toprule
\multirow{2}{*}{\textbf{Variants}} & \multicolumn{3}{c|}{\textbf{Non-Edited Region}} & \multicolumn{4}{c}{\textbf{Edited Region}} \\
\cmidrule(lr){2-4} \cmidrule(lr){5-8}
 & \textbf{PSNR}\(\uparrow\) & \textbf{SSIM}\(\uparrow\) & \textbf{LPIPS}\(\downarrow\) & \textbf{ACC}\(\uparrow\) & \textbf{NED}\(\uparrow\) & \textbf{CDD}\(\downarrow\) & \textbf{Harm.}\(\uparrow\) \\
\midrule
Reverse Direction & 27.679 & \underline{0.792} & \underline{0.116} & \underline{0.890} & \underline{0.979} & 0.121 & 1.551 \\
RePair w/o Filtering & \underline{28.329} & \textbf{0.803} & \underline{0.116} & \underline{0.890} & 0.978 & \underline{0.067} & \underline{2.606} \\
\midrule
\textbf{RePair (Ours)} & \textbf{28.443} & \textbf{0.803} & \textbf{0.114} & \textbf{0.916} & \textbf{0.981} & \textbf{0.063} & \textbf{2.761} \\
\bottomrule
\end{tabular}}
\end{table}

\begin{table}[t]
\centering
\caption{Ablation study on Region-Decoupling Optimization (RDO) and Cascaded High-Frequency Texture Guidance (CHFTG). RDO applies distinct optimization targets to edited and non-edited regions. When \textit{RDO} is disabled, \textit{CHFTG} is applied to both non-edited and edited regions. All reported scores are evaluated on the HarmonyText-Bench.}
\label{tab:ablation_spatial_freq}
\renewcommand{\arraystretch}{0.90}
\resizebox{\linewidth}{!}{
\begin{tabular}{cc|ccc|cccc}
\toprule
\multirow{2}{*}{\textbf{RDO}} & \multirow{2}{*}{\textbf{CHFTG}} & \multicolumn{3}{c|}{\textbf{Non-Edited Region}} & \multicolumn{4}{c}{\textbf{Edited Region}} \\
\cmidrule(lr){3-5} \cmidrule(lr){6-9}
& & \textbf{PSNR}\(\uparrow\) & \textbf{SSIM}\(\uparrow\) & \textbf{LPIPS}\(\downarrow\) & \textbf{ACC}\(\uparrow\) & \textbf{NED}\(\uparrow\) & \textbf{CDD}\(\downarrow\) & \textbf{Harm.}\(\uparrow\) \\
\midrule
& & 27.658 & 0.798 & 0.120 & 0.903 & 0.979 & 0.145 & 0.952 \\
\(\checkmark\) & & \underline{28.401} & \underline{0.802} & \underline{0.115} & 0.890 & 0.979 & \underline{0.071} & \underline{2.457}  \\
& \(\checkmark\) & 27.703 & 0.798 & 0.120 & \textbf{0.923} & \textbf{0.986} & 0.137 & 1.063 \\
\midrule
\(\checkmark\) & \(\checkmark\) & \textbf{28.443} & \textbf{0.803} & \textbf{0.114} & \underline{0.916} & \underline{0.981} & \textbf{0.063} & \textbf{2.761} \\
\bottomrule
\end{tabular}
}
\end{table}

\begin{figure}[!t]
    \centering
    \includegraphics[width=\linewidth]{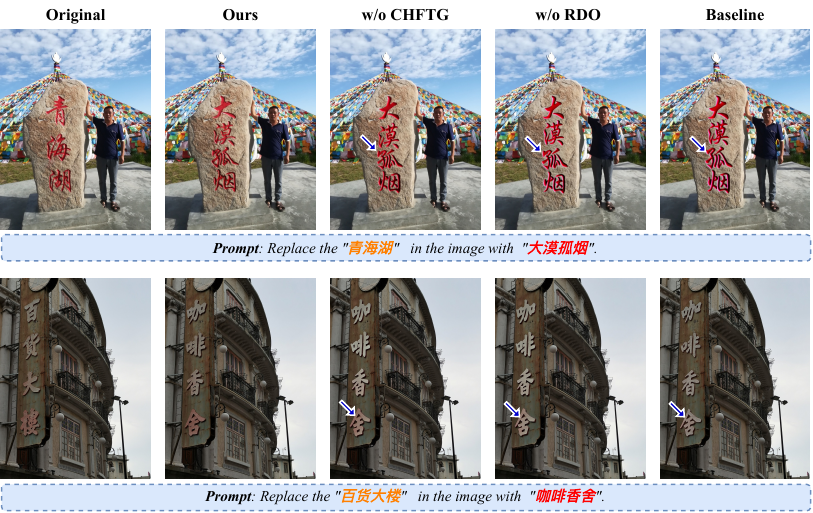}
    \caption{\textbf{Visual ablation on Region-Decoupled Frequency Learning.} \textit{w/o RDO} removes Region-Decoupling Optimization, applying a single optimization target to the entire image. \textit{w/o CHFTG} removes Cascaded High-Frequency Texture Guidance, relying solely on spatial-domain supervision. Arrows indicate regions where color or texture deviates from the original image. The full method yields the best overall consistency. \textbf{Best viewed with zoom-in.}}
    \Description{Visual ablation of Region-Decoupling Optimization and Cascaded High-Frequency Texture Guidance. The comparison shows outputs without RDO, without CHFTG, and with the full method, with arrows highlighting color and texture inconsistencies.}
    \label{fig:ablation_loss}
\end{figure}

\paragraph{Ablation on RDO and CHFTG}
Tab.~\ref{tab:ablation_spatial_freq} ablates Region-Decoupling Optimization (RDO) and Cascaded High-Frequency Texture Guidance (CHFTG). Without either component, the baseline yields the highest CDD (0.145) and the lowest Harmony score (0.952). RDO alone produces the largest gain, reducing CDD from 0.145 to 0.071 and raising the Harmony score from 0.952 to 2.457, confirming that source-anchored background supervision is the key driver of visual harmony. Notably, RDO slightly decreases ACC (0.903$\to$0.890), indicating that spatial decoupling alone lacks explicit guidance for sharp text stroke generation. CHFTG alone improves text accuracy (ACC 0.903$\to$0.923, NED 0.979$\to$0.986) through wavelet-domain constraints that sharpen text boundaries. However, it provides limited improvement in edited-text color consistency (CDD 0.145$\to$0.137), since global frequency supervision without region separation does not explicitly constrain the edited text to preserve the source color distribution. Combining both reveals a clear synergy. RDO's region-aware masks enable CHFTG to apply frequency constraints selectively, recovering ACC from 0.890 to 0.916 while further reducing CDD to 0.063 and raising the Harmony score to 2.761. As illustrated in Fig.~\ref{fig:ablation_loss}, all ablated variants exhibit color drift and texture inconsistency in the non-edited regions (indicated by blue arrows), confirming that both components are indispensable.

\subsection{Limitations}
HarmonyText depends on PaddleOCR-v5 for text localization, HiSAM for fine-grained text segmentation, and Qwen3-VL for instruction generation and verification. Errors from these components may propagate to data construction or evaluation. Although a bounding-box fallback keeps CDD well-defined when the source and edited text masks have an empty intersection, inaccurate localization can still reduce its precision. Curved, very small, and low-resolution text remains challenging, and the current evaluation mainly covers Chinese and English; the public-dataset results therefore provide broader evidence but do not establish universal linguistic generalization.

\section{Conclusion}

In this paper, we presented \textbf{HarmonyText}, a mask-free scene text editing framework that improves the reliability and realism of instruction-guided text modification in natural scenes. At its core, \textbf{Region-Decoupling Optimization} mitigates the background inconsistency in pseudo-target supervision by assigning different objectives to edited and preserved regions, allowing the model to generate target text while preserving original background details. \textbf{Cascaded High-Frequency Texture Guidance} further enhances this design by explicitly supervising wavelet-domain details, encouraging the learning of sharper text boundaries and more faithful local textures. Furthermore, we introduce Reverse-to-Refined Pair Synthesis to construct high-quality paired training data from imperfect edits, and design Color Distribution Divergence and the Harmony Assessment Model to enable GT-free evaluation of color consistency and perceptual text-background integration. Experiments show that HarmonyText produces more accurate, coherent, and visually harmonious edits than existing methods. Evaluation on HarmonyText-Bench and 345 additional public-source images, together with a blinded user study, further supports the effectiveness and generalization of the proposed framework, providing a solid foundation for future research on realistic scene text editing.

\begin{figure*}[!t]
    \centering
    \includegraphics[width=0.72\textwidth]{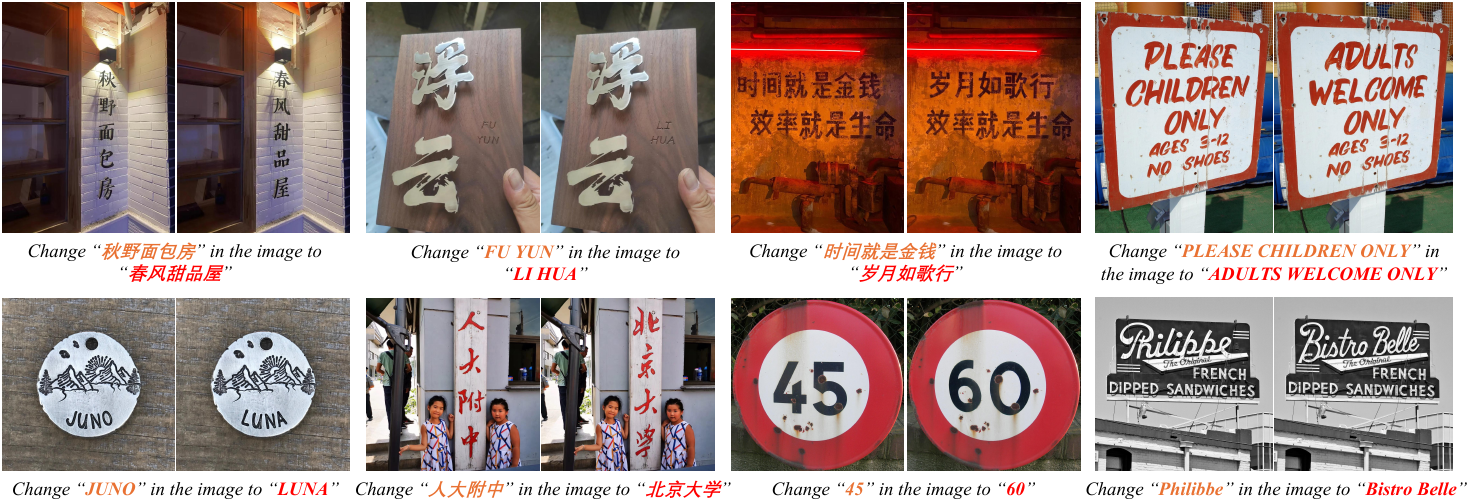}
    \caption{\textbf{Representative samples from the STE25K dataset.} Each pair shows the source image (left) and the edited result (right) along with the corresponding editing instruction. STE25K covers diverse real-world scenarios including both Chinese and English text across various surface materials, demonstrating the high quality and diversity of the paired data produced by our RePair pipeline.}
    \Description{Representative STE25K source-and-edited image pairs with text-editing instructions. The examples cover Chinese and English text on varied real-world surfaces and demonstrate diverse colors, textures, layouts, and backgrounds.}
    \label{fig:ste25k}
\end{figure*}

\begin{acks}

This work was supported in part by the
\grantsponsor{GS501100001809}{National Natural Science Foundation of China}
{https://doi.org/10.13039/501100001809}
under Grant No.~\grantnum{GS501100001809}{62476093};
the
\grantsponsor{GS501100003453}{Natural Science Foundation of Guangdong Province}
{https://doi.org/10.13039/501100003453}
under Grant No.~\grantnum{GS501100003453}{2026A1515012038};
the
\grantsponsor{GS501100002858}{China Postdoctoral Science Foundation}
{https://doi.org/10.13039/501100002858}
under Grant No.~\grantnum{GS501100002858}{2026M791625};
and the Postdoctoral Fellowship Program (Grade B) of the China Postdoctoral Science Foundation
under Grant No.~\grantnum{GS501100002858}{GZB20260386};
and by KlingAI Research.
\end{acks}

\bibliographystyle{ACM-Reference-Format}
\bibliography{ref}

\end{document}